\documentclass[runningheads]{llncs}

\usepackage{booktabs}
\usepackage{array}
\usepackage{atbegshi}
\usepackage{microtype}
\usepackage{xcolor}
\usepackage{hyperref}
\usepackage{xspace}
\usepackage{amsmath}
\usepackage{graphicx}
\usepackage{tikz}
\usetikzlibrary{arrows.meta,positioning,fit}

\hypersetup{colorlinks=true,linkcolor=blue,citecolor=blue,urlcolor=blue}
\newcommand{\system}{\textsc{Skill Runtime Intelligence}\xspace}
\newcommand{\panorama}{\textsc{Run Panorama}\xspace}
\newcommand{\observed}{\textsc{Observed}\xspace}
\newcommand{\derived}{\textsc{Derived}\xspace}
\newcommand{\inferred}{\textsc{Inferred}\xspace}
\newcommand{\experimental}{\textsc{Experimental}\xspace}

\title{Evidence-Calibrated Runtime Reconstruction for Agent Skills Across
Heterogeneous Coding Agents}
\titlerunning{Evidence-Calibrated Runtime Reconstruction for Agent Skills}
\author{Xueping Gao}
\authorrunning{X. Gao}
\institute{Alibaba Cloud, Hangzhou, China, \email{xueping.gxp@alibaba-inc.com}}

\ifdefined\XeTeXversion
  
  \AtBeginShipout{}
\fi

\begin{document}
\maketitle

\begin{abstract}
Agent Skills package reusable instructions and assets for tool-using
language-model agents. Progressive loading creates failure boundaries poorly
represented by session-, model-, or tool-centric traces: a Skill can be
discovered but not activated, activated without instructions, or appear
successful without an independently verified outcome. We present \system, a
passive runtime-intelligence system that reconstructs supported Skill-lifecycle
stages across heterogeneous harnesses while preserving unsupported stages as
unknown. Its \panorama separates immutable events, deterministic relations,
inferred diagnoses, and controlled outcomes with four evidence grades; optional
trace import and OTLP/HTTP export support existing observability deployments.

Across six frozen repository profiles, three coding agents, and seven clean or
fault-injected conditions, all 126 executions preserve source worktrees and
each correlates to exactly one source session. Yet adapters expose three distinct semantics:
no Skill runs; complete runs but no failure-like events; or failure-like events
in every operational-failure and clean session. In a seven-template diagnostic
study, semantic aliases and Panorama localize the same six non-clean boundaries
but differ in exact/status behavior; both Raw views emit a failure status on all
18 clean cases, while Panorama emits none. A known-rule graph conforms to
126/126 frozen contracts, whereas a second model completes only 228/378 calls.
These observations motivate executable adapter qualification and show that event
presence is not boundary fidelity, composite exact scores mask distinct errors,
and model explanations must not overwrite deterministic facts.
\end{abstract}

\keywords{Agent Skills \and runtime observability \and evidence provenance
\and coding agents \and empirical software engineering}

\section{Introduction}

Language-model agents increasingly acquire reusable capabilities through
\emph{Skills}: directories containing declarative metadata and optional
scripts, references, and assets.

The specification uses progressive disclosure: lightweight metadata may
be visible before full instructions or resources are loaded
\cite{agent-skills-spec}.  This improves modularity but creates a new operational
question: \emph{did the Skill actually run as intended?}

The final answer is insufficient evidence.  A plausible answer may follow a
missed activation, a missing reference, a failed command, or an unverified
artifact.  Conversely, missing telemetry does not prove that a lifecycle step
failed.  Existing agent evaluation commonly measures task success or trajectory
quality \cite{swebench,agentdiagnose}; general tracing standards model
agent and tool spans \cite{otel-genai}, but do not make a dynamically loaded
Skill occurrence, its resource boundary, or its evidence grade the primary
entity.  The operational gap is therefore not simply ``more tracing.''  It is
the reconstruction of a portable Skill lifecycle from incomplete,
harness-specific evidence without inventing observations.

We introduce \system, a passive runtime evidence architecture for Agent Skills.
It observes existing agent workflows rather than proxying model requests or
orchestrating the agent.  Versioned adapters retain raw source events and map
them into a common event model.  A deterministic evidence graph reconstructs
Skill occurrences and lifecycle relations.  Diagnoses state both their
evidential basis and their causal scope.  Model-based analysis is restricted to
an \inferred layer and cannot promote or overwrite deterministic facts.

Our study asks four research questions:

\begin{description}
  \item[RQ1:] How faithfully can heterogeneous agent telemetry reconstruct a
  Skill lifecycle and its first divergent boundary?
  \item[RQ2:] How do exact diagnosis, boundary localization, status, and
  evidence entailment differ between a normalized Panorama and raw-event views,
  and when are deterministic or model-assisted diagnoses justified?
  \item[RQ3:] Can collection remain non-intervening, silent, and low-overhead
  in the tested deployment environments?
  \item[RQ4:] What controlled mechanism coverage is observed across frozen
  repository profiles, agents, environments, and model backends?
\end{description}

This paper makes four contributions:

\begin{enumerate}
  \item A Skill-specific runtime model that distinguishes occurrence from
  relationship attribution.
  \item A four-grade evidence contract---\observed, \derived, \inferred, and
  \experimental---with explicit causal-scope restrictions.
  \item A deployment-flexible system with versioned adapters, immutable
  evidence, deterministic reconstruction, and non-authoritative model
  assistance.
  \item A controlled benchmark of six frozen repository profiles, three agents,
  seven condition families, two model backends, outcome verification, and
  template-level comparisons.
\end{enumerate}

As a process contribution, the frozen matrix qualifies each adapter--agent
release: calibrate capabilities, publish limits, and rerun the gates. The study
does not estimate natural incident prevalence, human usability, causal Skill
effectiveness, or unconstrained production diagnosis accuracy.

\section{Problem and Design Requirements}

\subsection{The Skill lifecycle}

We represent one attempted Skill execution as the ordered lifecycle

\begin{center}
\small
Request $\rightarrow$ Discovery $\rightarrow$ Activation $\rightarrow$
Instructions $\rightarrow$ Resources $\rightarrow$ Execution $\rightarrow$
Artifacts $\rightarrow$ Outcome.
\end{center}

The stages are logical boundaries, not a requirement that every harness emit a
span for every stage.  A resource access can be directly observed while its
membership in a particular Skill occurrence is only deterministically derived.
An agent may report success while the outcome remains unverified.  This
separation prevents two common errors: treating absence of telemetry as a
failure, and treating an agent assertion as an independently verified result.

\subsection{Requirements}

Our design follows six requirements.

\paragraph{R1: Observation without takeover.}
The system must not proxy model requests, own the agent loop, or block actions
by default.  Collection should be compatible with normal agent use.

\paragraph{R2: Source preservation.}
Raw records are immutable and remain separately addressable.  Normalization
cannot erase source identity, and multiple physical streams with the same
upstream session identifier cannot overwrite one another.

\paragraph{R3: Epistemic separation.}
Facts directly supplied by a source, deterministic transformations, uncertain
analysis, and controlled effect estimates must not share one undifferentiated
confidence field.  This follows the broader provenance principle that entities,
activities, and derivations should retain explicit relations
\cite{w3c-prov-dm}.

\paragraph{R4: Versioned capability.}
Adapters are measurement instruments whose schemas and coverage change with
agent versions.  Every normalized record binds agent, agent version, adapter
version, and source format.  Unsupported fields remain unknown.

\paragraph{R5: Privacy by minimization.}
Prompts, code, paths, credentials, and raw payloads are not required for most
lifecycle diagnoses. Evidence remains within the operator-controlled deployment
boundary---local by default or an explicitly authenticated self-hosted
service---and exported research views contain only minimum ordered states and
opaque evidence identifiers.

\paragraph{R6: Causal restraint.}
A single run may establish that an event occurred or that a verifier passed; it
cannot establish that the Skill caused the outcome.  Skill-effect claims require
controlled repeated trials and are labeled \experimental.

\section{System Architecture}

\begin{figure}[t]
\centering
\begin{tikzpicture}[
  font=\scriptsize,
  node distance=3mm and 3mm,
  box/.style={draw,rounded corners,align=center,minimum height=11mm,
    text width=19mm,inner sep=1mm,fill=blue!4},
  oracle/.style={box,fill=orange!10},
  model/.style={box,fill=green!8},
  arr/.style={-{Latex[length=1.7mm]},semithick},
  dasharr/.style={arr,dashed}
]
\node[box] (agents) {Existing\\agent runs};
\node[box,right=of agents] (raw) {Immutable\\raw records};
\node[box,right=of raw] (adapter) {Versioned\\normalized events};
\node[box,right=of adapter] (graph) {Evidence graph\\and Panorama};
\node[box,right=of graph] (finding) {Evidence-citing\\findings};
\draw[arr] (agents)--(raw);
\draw[arr] (raw)--(adapter);
\draw[arr] (adapter)--(graph);
\draw[arr] (graph)--(finding);

\node[oracle,below=5mm of agents] (manifest) {Frozen fault\\manifest + verifier};
\node[oracle,right=of manifest] (gold) {Frozen gold boundary,\\status, outcome};
\node[oracle,right=of gold] (score) {Offline scorer\\(evaluation only)};
\draw[arr] (manifest)--(gold);
\draw[arr] (gold)--(score);

\node[model,below=5mm of graph] (views) {Raw, Panorama,\\known-rule views};
\node[model,right=of views] (llm) {Optional model\\candidate\\(\inferred)};
\draw[arr] (graph)--(views);
\draw[arr] (views)--(llm);
\draw[dasharr] (llm)--(finding);
\draw[dasharr] (finding.east)--++(3mm,0)
  |- ([yshift=-3mm]score.south)--(score.south);
\end{tikzpicture}
\caption{Production reconstruction (blue), non-authoritative model candidates
(green), and evaluation-only oracle data (orange).  The offline scorer consumes
findings and frozen gold; it never backfills observed telemetry.}
\label{fig:architecture}
\end{figure}
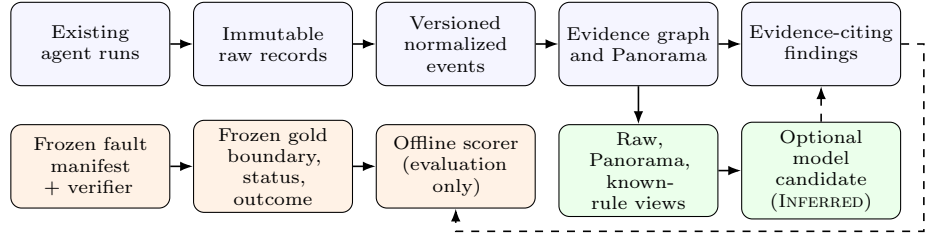

\subsection{Data path}

\system has four layers (Fig.~\ref{fig:architecture}):

\begin{enumerate}
  \item \textbf{Collectors} ingest official hooks, local transcripts, and
  read-only filesystem or process signals.
  \item \textbf{Versioned adapters} retain raw source records and emit a common
  event envelope with timestamp and source provenance.
  \item \textbf{The evidence engine} constructs sessions, turns, Skill runs,
  resources, tools, artifacts, outcomes, and typed relations.
  \item \textbf{The Panorama and diagnostics} expose the lifecycle, the first
  observable divergence, the supporting evidence, and unobservable boundaries.
\end{enumerate}

Deployment placement and observability interoperability are orthogonal. The
implementation runs locally or as an authenticated self-hosted service with
separate viewer and Collector credentials; trace import and OTLP/HTTP export
remain optional. OpenTelemetry's generative-AI conventions describe agent and
tool spans \cite{otel-genai}; \system adds a domain layer centered on a
versioned Skill definition and one occurrence of that definition.

\subsection{Event and identity model}

Each normalized event contains a locally unique identifier, event type,
occurrence and ingestion times, physical source session, optional turn and Skill
run identifiers, source locator, adapter identity, and evidence metadata.  The
event vocabulary covers session and turn state, Skill discovery and activation,
instruction and resource access, tool and subagent execution, workspace and
artifact activity, and reported, verified, or unknown outcomes.

Physical evidence streams and logical upstream session IDs are intentionally
separate.  A correlation key may join streams for analysis but never authorizes
destructive merging.  Similarly, an event's existence and the edge assigning it
to a Skill run are separate claims.  Exact path and identifier boundaries avoid
prefix collisions such as assigning \texttt{pdf-backup} activity to
\texttt{pdf}.

Stable identities combine adapter version, physical source-instance identity,
and explicit source event or call identifiers; timestamps alone never create
identity. Reconstruction applies fixed precedence: source parent/child ID,
explicit Skill attribution, active Skill scope, exact artifact path, temporal
adjacency, then model or heuristic suggestion. The first four may create
\derived edges. The last two remain uncertain and cannot replace a
higher-priority edge. Conflicting equal-priority relations are retained as
ambiguous rather than resolved by timestamp proximity.

The engine (1) appends the immutable raw record, (2) consults the
adapter-version capability declaration, (3) emits only supported normalized
fields, (4) attaches deterministic edges in precedence order, (5) traverses the
eight lifecycle stages, and (6) emits a finding only when an observed failure or
an evaluable expected signal establishes a boundary. An optional model runs
after this procedure. For example, a source \texttt{Skill} call is retained,
normalized to \texttt{skill.\allowbreak activated} as \observed, and connected to subsequent
tool activity by a separately graded \texttt{skill\_\allowbreak scope} edge; a model
explanation is stored in a different \inferred record. The implementation is
approximately 12.9k Python lines spanning versioned adapters, storage,
redaction, reconstruction, diagnostics, export, authenticated local/remote
service, and UI.

\subsection{Evidence grades and causal scope}

Table~\ref{tab:evidence} defines the evidence contract.  Grade is about how a
claim is known; causal scope is a separate authorization decision.  For example,
an independently verified subprocess failure is \observed or \derived evidence
of an outcome, but still does not establish a Skill-to-outcome causal effect.

\begin{table}[t]
\centering
\caption{Evidence grades. Cross-run causal effects require controlled trials.}
\label{tab:evidence}
\small
\begin{tabular}{@{}>{\raggedright\arraybackslash}p{0.18\columnwidth}%
  >{\raggedright\arraybackslash}p{0.70\columnwidth}@{}}
\toprule
Grade & Meaning \\
\midrule
Observed & Directly present in a source record or external verifier. \\
Derived & Deterministic transformation or relation over observed records. \\
Inferred & Uncertain model or heuristic explanation; never promoted silently. \\
Experimental & Estimate from a declared controlled experiment. \\
\bottomrule
\end{tabular}
\end{table}

\subsection{Deterministic diagnosis graph}

The evidence engine maps each lifecycle stage to observed success, observed
failure, expected but not observed, unsupported, or not applicable.
Versioned rules traverse ordered stages and typed edges to produce a candidate
boundary and status.  Each finding cites the exact node or relation that entails
it.  Missing events only become a finding when the adapter capability and an
independent expectation make the absence evaluable.

Model analysis receives a minimized view and returns a structured candidate,
status, evidence IDs, and causal flag.  It is stored as \inferred.  A combined
graph-and-model condition may explain or rank deterministic candidates, but the
production contract does not allow it to overwrite a graph fact.  The
experiment deliberately permits that change so the risk is measurable.

\section{Evaluation Methodology}

\subsection{Experimental principles}

We preregister the condition matrix and gates in machine-readable manifests.
Source execution, telemetry reconstruction, agent response, and model diagnosis
are separate endpoints.  Failed gates and malformed outputs remain in the
reports; we do not silently retry them.  Repositories are read through frozen
Git objects, so pre-existing dirty files are provenance covariates but not
experimental inputs.  SHA-256 binds fixtures, probes, and reports.

Our methodology resembles executable repository benchmarks such as SWE-bench
\cite{swebench} in freezing code state and using an external oracle, but differs
in target: we measure Skill-lifecycle evidence rather than patch correctness.

\subsection{Multi-repository Agent benchmark}

Table~\ref{tab:profiles} summarizes six three-file profiles: two Go, two Python
research, and two Agent Skill or specification projects.  Each receives a
repository-specific, read-only audit Skill.  A deterministic probe operates on
a temporary overlay and emits a nonce-bound oracle record; source worktrees
remain byte-identical.

\begin{table}[t]
\centering
\caption{Frozen repository profiles. Public aliases minimize identity
disclosure; revisions and files are manifest-bound; diversity is profile-level,
not repository-scale.}
\label{tab:profiles}
\scriptsize
\begin{tabular}{@{}lll@{}}
\toprule
Profile & Type & Frozen files (three each) \\
\midrule
profile-01 & Go cache & module config, implementation, test \\
profile-02 & Go telemetry & README, module config, agent \\
profile-03 & Python research & packaging, configuration, activation \\
profile-04 & Python app & README, dependencies, application \\
profile-05 & Spec workflow & specification, helper script, check \\
profile-06 & Skill catalog & README, Skill definition, helper script \\
\bottomrule
\end{tabular}
\end{table}

The benchmark crosses three installed coding-agent CLIs---Codex, OpenCode, and
Qoder---with seven conditions: clean, instruction failure, missing resource,
execution failure, artifact corruption, unverified outcome, and verifier
conflict.  This yields $6\times3\times7=126$ cells.  There is one execution per
cell; the experiment establishes mechanism coverage, not stochastic population
rates.  Per agent, the matrix contains 24 operational failure cells (four
failure boundaries across six repositories), six clean cells, and 12
outcome-evidence cells (unverified outcome and verifier conflict).  We report
these denominators separately because outcome-evidence conflicts are not native
execution failures.

Faults target Instructions, Resources, Execution, Artifacts, and Outcome.
Request occurs in every cell; Discovery and Activation appear on successful
paths but are not fault-injected. Thus five of eight lifecycle stages are
evaluated as fault locations.

We use two independent gates.  The \emph{integrity gate} requires every cell to
launch, preserve its workload, and correlate to exactly one collected source
session.  The stricter \emph{response gate} additionally requires the agent to
return the exact nonce-bound oracle JSON.  Reconstruction is scored separately
for Skill-run coverage, failure-like-event co-occurrence, clean
failure-like-event rate, and exact first-boundary localization.  The
co-occurrence metric does not attribute an arbitrary failed event to the
injected nonce or boundary.

Table~\ref{tab:agents} fixes the execution interfaces used in this study.  These
are adapter--version observations, not rankings of the underlying models or
products.

\begin{table}[t]
\centering
\caption{Installed agent interfaces used for the 126-cell benchmark.}
\label{tab:agents}
\small
\begin{tabular}{@{}llll@{}}
\toprule
CLI & Version & Model/mode & Environment \\
\midrule
Codex & 0.145.0 & \texttt{gpt-5.6-sol} & macOS arm64 \\
OpenCode & 1.18.5 & \texttt{deepseek-v4-flash-free} & macOS arm64 \\
Qoder & 1.1.10 & \texttt{performance} & macOS arm64 \\
\bottomrule
\end{tabular}
\end{table}

\subsection{Diagnostic utility study}

The 126 collected cases are transformed into a privacy-safe holdout that omits
raw content, paths, payloads, timestamps, and source session IDs.  Gold labels
come from the frozen fault manifest and deterministic probe, not a model or
human adjudicator.  The cases repeat seven semantic condition templates across
repositories and agents; they are paired cases, not 126 independent fault
families.  Raw and Panorama are programmatic views of the same controlled
execution.  We compare four primary views and one Qwen-only control:

\begin{enumerate}
  \item \textbf{Raw model}: ordinal native events with minimal structure;
  \item \textbf{Semantics-matched Raw}: every native and auxiliary record is
  retained while inline aliases expose Panorama-equivalent lifecycle, kind,
  status, and evidence-grade semantics;
  \item \textbf{Panorama model}: normalized lifecycle stages and evidence grades;
  \item \textbf{Known-rule graph}: deterministic typed-relation candidate;
  \item \textbf{Graph+Model}: the graph candidate plus a model explanation pass.
\end{enumerate}

For the primary model conditions, each case is evaluated in Raw, Panorama, and
Graph+Model views, or 378 calls per backend. The semantics-matched control adds
126 Qwen calls. Primary metrics are exact boundary-and-status diagnosis,
boundary accuracy, status accuracy, citation-ID validity, citation entailment,
causal safety, completion, and latency.  Citation entailment is stricter than
checking whether an identifier exists: the cited record must support the
predicted relation. We report paired case counts as descriptions of the frozen
matrix and compare higher/lower/equal directions at the seven-template level.
Rows within a condition template are strongly dependent, so we report no
case-level significance test or population inference.

We run Qwen3.6-35B-A3B through vLLM on a remote PAI-DSW Linux x86\_64 instance.
We independently request the same matrix from DeepSeek-v4 through the installed
OpenCode interface on macOS.  A backend passes only if all requested calls
complete with valid structured citations and causal safety.  The Qwen endpoint
uses temperature 0, 384 maximum output tokens, thinking disabled, and a strict
JSON schema; the OpenCode path fixes CLI and model versions but does not expose
equivalent service-side decoding controls.

Two secondary stress tests probe the boundary of this comparison.  The first
selects 19 de-identified runtime traces from one local Panorama database, with
at most four cases per deterministic finding profile.  The labels are production
rule candidates, not human gold.  The second balances six preregistered
rule-external graph anomalies with six clean controls; these cases test guarded
hypothesis generation rather than natural incident prevalence.

For an additional real-trace adjudication, we remove every deterministic
candidate and label-origin field before annotation. Qwen and Codex independently
see only the 19 de-identified evidence graphs and a frozen category rubric. The
rule candidates are revealed only after both reports are stored. This is a
blinded model-adjudication check, not human ground truth; disagreement and
abstention are retained.

\subsection{Non-intervention and reproducibility}

Collector microbenchmarks pair an instrumented call with an uninstrumented
control and verify input/output equivalence, exact delivery, silent success,
and failure isolation.  A remote Linux x86\_64 study executes both direct and
shell transport paths; a second Linux arm64 container rebuilds and executes the
native sender.  These environments test mechanism portability, not independent
physical-host reliability.

The repository's reproducibility suite contains 12 deterministic correctness
gates and one environment-sensitive transport gate.  Unit and integration
tests cover adapter identity, event normalization, reconstruction, diagnosis,
privacy export, and experimental report contracts.

\section{Results}

\subsection{RQ1: Reconstruction is harness-dependent}

All 126 Agent cells satisfy the integrity gate: source worktrees are unchanged,
and every call maps to exactly one source session.  The response gate fails:
122/126 responses match the oracle.  Codex and Qoder return 42/42; OpenCode
returns 38/42.  The four OpenCode failures remain in the analysis.

Table~\ref{tab:reconstruction} shows the central reconstruction result. Codex's
42/42 exact nonce-bound responses and target-\texttt{SKILL.md} path signatures
rule out complete task non-execution, yet its adapter reconstructs no Skill
runs; neither signal establishes hidden activation semantics. OpenCode achieves
full Skill-run coverage but emits no failure-like events in the 24 operational
failure cells.  Qoder emits at least one failure-like event in all 24 failure
cells, but also in all six clean cells, and exactly localizes six of the 24
injected boundaries.  Because the event is not nonce-attributed, 24/24 is a
session-level co-occurrence count rather than injected-failure detection.  Event
presence therefore cannot be interpreted as faithful boundary semantics.

\begin{table}[t]
\centering
\caption{Reconstruction counts over 42 sessions per agent. Failure-like-event
denominators are 24 operational failure cells and six clean cells.}
\label{tab:reconstruction}
\small
\begin{tabular}{@{}l@{\hspace{10pt}}r@{\hspace{10pt}}r@{\hspace{10pt}}r@{\hspace{10pt}}r@{}}
\toprule
Agent & Skill runs & Failure & Clean & Exact bnd. \\
\midrule
Codex & 0/42 & 0/24 & 0/6 & 0/24 \\
OpenCode & 42/42 & 0/24 & 0/6 & 0/24 \\
Qoder & 42/42 & 24/24 & 6/6 & 6/24 \\
\bottomrule
\end{tabular}
\end{table}

This is not evidence that one agent is intrinsically more reliable.  It is
evidence that the current versioned adapters expose different measurement
capabilities.  Product reporting must combine coverage, clean specificity,
attribution, and localization rather than displaying a binary ``failure events
supported'' badge.

\subsection{RQ2: Semantics and structure show different diagnostic signatures}

Table~\ref{tab:qwen} reports the complete Qwen study. Panorama has 10 more exact
diagnoses and 36 more correct boundaries than Raw. The stricter
semantics-matched Raw control also localizes 108 boundaries, consistent with
named lifecycle aliases being sufficient for the matched boundary count under
this frozen prompt contract. Yet it
has only 49 exact diagnoses and 49 correct statuses; compact Panorama reaches 82
and 100. Thus the two views have different template-level signatures for status
and exactness even when boundary localization matches. Citation entailment
remains non-monotonic.

The known-rule graph reproduces all 126 contract-generated labels. This is
expected conformance because graph rules and gold labels share the frozen fault
contract; it is not independent diagnostic accuracy. Graph+Model is exact on
125. Its sole mismatch is a clean case with correct
\texttt{verified\_success} status but boundary \texttt{outcome} rather than the
gold convention \texttt{none}. This is a boundary-label sensitivity
observation, not evidence that model augmentation generally reduces accuracy.
Citation entailment is only 89/126 in Graph+Model despite 125 exact answers.

\begin{table}[t]
\centering
\caption{Qwen results (126 cases per view). ``Ent.'' counts supported cited
relations; ``Clean-FP'' counts failure statuses on 18 clean controls. The last
two rows are contract references, not independent model baselines.}
\label{tab:qwen}
\small
\begin{tabular}{@{}l@{\hspace{8pt}}r@{\hspace{8pt}}r@{\hspace{8pt}}r@{\hspace{8pt}}r@{\hspace{8pt}}r@{}}
\toprule
View & Exact & Bnd. & Status & Ent. & Clean-FP \\
\midrule
Raw model & 72 & 72 & 99 & 62 & 18/18 \\
Semantics-matched Raw & 49 & 108 & 49 & 26 & 18/18 \\
Panorama model & 82 & 108 & 100 & 46 & 0/18 \\
\midrule
Known-rule graph & 126 & 126 & 126 & 126 & 0/18 \\
Graph+Model & 125 & 125 & 126 & 89 & 0/18 \\
\bottomrule
\end{tabular}
\end{table}

Qwen completes the 378 primary calls and 126 follow-up control calls; median
latency is 2.16--2.35 seconds per view. The causal-safety and
citation-ID-validity gates pass for all model views. A retained prompt-legend
pilot reaches only 26/126 exact, motivating the inline-alias control and showing
that status predictions are sensitive to seemingly equivalent natural-language
scaffolding under this prompt contract. These interface gates do not imply
diagnostic correctness.

Table~\ref{tab:template-results}(a) compares the stricter control with Panorama without
treating the 126 rows as independent samples. Boundary localization is equal in
all seven templates. Panorama has a higher exact count in three templates,
Semantics-matched Raw in one, and they are equal in three. Status and entailment
directions also vary, so the case totals are descriptive rather than evidence of
a population effect.

Condition-family strata in Table~\ref{tab:template-results}(b) show why the exact count
must not be summarized as a uniform benefit. Panorama has more correct counts
for artifact corruption and verifier conflict, but fewer exact diagnoses for
execution and instruction failures. Its 26 status errors are a single
directional confusion:
\texttt{observed\_failure} is predicted as \texttt{verifier\_conflict} in all
18 instruction cases and eight execution cases. This observed pattern could
arise from the interface or model and is not attributed without a preregistered
follow-up.

\begin{table}[t]
\centering
\caption{Template-aware Qwen comparisons: (a) S-Raw and Panorama totals and
higher/equal templates; (b) Raw:Panorama correct counts per 18-instance
template. All counts are descriptive.}
\label{tab:template-results}
\scriptsize
\begin{minipage}[t]{0.39\textwidth}
\centering
\textbf{(a) S-Raw versus Panorama}\\[2pt]
\begin{tabular}{@{}lrrrr@{}}
\toprule
Metric & S/P & P+ & S+ & Eq. \\
\midrule
Exact & 49/82 & 3 & 1 & 3 \\
Boundary & 108/108 & 0 & 0 & 7 \\
Status & 49/100 & 4 & 1 & 2 \\
Entailment & 26/46 & 3 & 2 & 2 \\
\bottomrule
\end{tabular}
\end{minipage}\hfill
\begin{minipage}[t]{0.59\textwidth}
\centering
\textbf{(b) Raw versus Panorama}\\[2pt]
\begin{tabular}{@{}l@{\hspace{10pt}}r@{\hspace{10pt}}r@{\hspace{10pt}}r@{}}
\toprule
Condition & Exact & Boundary & Status \\
\midrule
Artifact corruption & 0:18 & 0:18 & 18:18 \\
Clean & 0:0 & 0:0 & 0:18 \\
Execution failure & 18:10 & 18:18 & 18:10 \\
Instruction failure & 18:0 & 18:18 & 18:0 \\
Outcome unverified & 18:18 & 18:18 & 18:18 \\
Resource missing & 18:18 & 18:18 & 18:18 \\
Verifier conflict & 0:18 & 0:18 & 9:18 \\
\bottomrule
\end{tabular}
\end{minipage}
\end{table}

The clean row exposes a limitation of the conjunctive Exact metric. Raw and
Semantics-matched Raw predict a failure status on all 18 clean instantiations.
Panorama predicts \texttt{verified\_success} on all 18, hence zero
failure-state false positives, but remains 0/18 Exact because it emits boundary
\texttt{outcome} rather than the frozen gold convention \texttt{none}. These
are different error types; neither count estimates a production false-positive
rate.

\subsection{Model availability is part of utility}

The independent DeepSeek matrix completes only 228/378 calls.  Of 150 failures,
111 time out and 39 violate the structured-output contract.  Completed subsets
are accurate---69/72 Raw, 72/78 Panorama, and 76/78 Graph+Model---but these
conditional values do not estimate full-matrix reliability.  Median latency for
completed views ranges from 26.9 to 30.1 seconds.  The preregistered completeness
and safety gate fails.

This negative result changes the product design: deterministic graph diagnosis
must remain available under timeout, malformed output, or provider degradation.
A model can augment an answer, but cannot be on the critical path to a
reproducible baseline diagnosis.

\subsection{Secondary stress tests bound the model role}

On the 19 de-identified runtime traces, an instruction-rich Qwen pass supplies
existing citation IDs in 19/19 stored outputs but relation-specific rescoring
finds entailment in 0/19. A DeepSeek pass exactly reproduces the deterministic
finding set in 5/19 and entails citations in 3/19. In the stricter blinded v2
adjudication, Qwen and Codex both complete 19/19 with valid citation IDs and
causal restraint. Their complete finding sets agree in 11/19; every one of these
11 strict consensus sets matches the hidden deterministic candidate. Individually,
Qwen and Codex match 11/19 and 19/19, while Qwen follows the one-finding-per-code
rubric in only 15/19. Because both annotators share the ontology and there is no
human ground truth, this supports the rule candidates only on a consensus subset
and exposes surplus-finding fragility; it is not real-world accuracy.

On the six rule-external anomaly/control pairs, Qwen and DeepSeek complete all
12 cases with precision/recall $1.00/.67$ and $.86/1.00$ ($F_1=.800/.923$):
their errors favor precision and recall, respectively, within this controlled
set. Cited support validates in 10/12 and 11/12 cases; predictions and support
are jointly valid in 9/12, leaving three disagreements. The production-rule baseline detects zero by
construction because these families are held outside its rule set.  These
controlled results motivate a narrow positive role for models: propose
\inferred candidates beyond existing rules, validate cited node relations, and
promote a reviewed recurring pattern into a versioned rule.  They do not measure
unknown-fault accuracy in production.

\subsection{RQ3: Passive collection can preserve execution}

Across the final 126-cell matrix, workload mutations are zero.  In a separate
balanced 60-call outcome study, all calls preserve the workload and an external
verifier confirms every success or non-zero subprocess failure.  Yet none of
the exact-matched sessions from the three adapters contains an explicit
normalized failure event.  Verified outcome and runtime lifecycle evidence are
therefore two lanes, neither of which may be fabricated from the other.

On remote Linux x86\_64, five default hook-transport runs deliver 400/400 events
exactly. Direct and shell incremental p95 overheads are 0.706 and 1.275 ms
(actual p95 1.163 and 1.739 ms), respectively; all calls are silent with zero
exit failures. A second Linux arm64 environment delivers 80/80 events, with
direct and shell incremental p95 of 2.354 and 1.871 ms (actual p95 2.677 and
2.842 ms).
These observations support low-overhead mechanism execution in the tested
environments, not a universal end-to-end latency or reliability bound.

\subsection{RQ4: Mechanism coverage across controlled profiles}

The reconstruction gap appears across six frozen repository profiles rather
than a single synthetic repository.  The full Qwen result also spans all three
agent sources and seven conditions. Together, these support controlled
mechanism coverage across repository profile, installed agent, fault boundary,
and one independently hosted model backend.

The external-validity boundary remains important.  Primary fault overlays are
controlled and oracle-backed rather than naturally occurring incidents.  There
is one execution per Agent--repository--condition cell.  The 19-trace stress test
comes from one local database and uses deterministic candidates rather than
independent human judgments.  We therefore do not claim incident prevalence,
human diagnostic benefit, or unrestricted semantic diagnosis.

\section{Discussion}

\subsection{Event presence is not boundary fidelity}

The three adapters occupy qualitatively different failure modes: missing Skill
occurrences, reconstructed occurrences without failure semantics, and ubiquitous
failure-like events with no clean specificity.  A single coverage percentage
would conceal all three.  Adapter capability should be calibrated per version
using a matrix of lifecycle coverage, failure-event co-occurrence, clean
specificity, attribution, and exact-boundary accuracy. Unknown versions should begin as unsupported rather
than inheriting historical capabilities.

\subsection{Answer and explanation quality are orthogonal}

The named-alias view has 108 correct Qwen boundary localizations versus 72 for
minimally structured Raw; compact normalization has the same boundary count but
different exact-status and entailment counts. Even Graph+Model's 125/126 exact
predictions have only 89/126 entailed citations. Interfaces should display answer
status, citation validity, relation entailment, and evidence grade separately.

\subsection{Deterministic core, probabilistic edge}

Our results suggest a division of labor.  Versioned rules should own known,
formalizable lifecycle relations.  Models can summarize them, rank review
items, or propose novel \inferred candidates.  A proposed pattern that survives
review and deterministic support checks should become a versioned rule rather
than remain permanently model-dependent.  This differs from trajectory
diagnosis systems that use an LLM judge as the final localizer
\cite{agentrx}; our architecture preserves a useful baseline even when the
judge is unavailable.

\subsection{Outcome and telemetry require two lanes}

An external test can verify a failure even when the harness emits no native
failure event.  Conversely, a native failure-like event may be present during a
clean execution.  The Panorama should show \emph{External Outcome} and
\emph{Runtime Evidence} side by side.  An outcome must not be backfilled as an
\observed lifecycle event, and missing lifecycle evidence must not erase a
verified outcome.

\subsection{Design implications for development workflows}

The following are design implications, not measured process-effect claims. The
architecture suggests changes to three recurring development activities.
First, every released adapter--agent version pair runs the lifecycle matrix and
publishes its coverage,
clean specificity, attribution, and boundary profile.  An untested version starts
as unsupported instead of inheriting an older capability badge.  Second, incident
triage follows an evidence-first sequence: preserve raw records, locate the first
observable divergence, compare the separate outcome lane, then request
a model explanation only for unresolved relations.  Third, a reviewed inferred
pattern graduates into a versioned deterministic rule with a regression fixture.

For a Skill author, the resulting loop is concrete: reproduce the run, inspect
the Panorama boundary, open the cited raw record, distinguish missing telemetry
from verified failure, repair the Skill or adapter, and rerun the same frozen
probe. This defines a candidate maintenance and regression-diagnosis workflow
without placing a model provider on the critical path. The study does not establish reduced
human repair time; that is a future usability outcome rather than a result of
the present controlled corpus.

\section{Related Work}

\paragraph{Agent execution and evaluation.}
Agent benchmarks evaluate tool-using systems on increasingly realistic tasks;
SWE-bench, for example, grounds software-agent outcomes in executable
repository tests \cite{swebench}. These efforts evaluate agent capability; our
focus is reconstructing a dynamically loaded Skill without owning the harness.

\paragraph{Trajectory diagnosis.}
AgentDiagnose analyzes competencies and visual patterns in agent trajectories
\cite{agentdiagnose}.  AgentRx localizes critical failure steps by synthesizing
constraints and applying an LLM judge to a validation log \cite{agentrx}.
HarnessFix compiles traces and harness code into a harness-aware IR, attributes
responsible steps and layers, and generates scoped repairs \cite{harnessfix}.
AgentDebugX closes the loop from observability and attribution to recovery and
rerun, including an installable debugging Skill \cite{agentdebugx}.
TraceElephant contrasts full and partial traces for multi-agent failure
attribution, while HarnessAudit evaluates full trajectories for boundary
compliance and execution fidelity \cite{traceelephant,harnessaudit}. These
systems diagnose, audit, or repair task- and harness-level trajectories. Our unit
is one progressively loaded Skill occurrence under heterogeneous, incomplete
telemetry. \system is passive and deployment-flexible, does not repair or rerun
the agent, grades evidence separately from causal scope, and prevents model
output from replacing deterministic relations.

\paragraph{Observability and provenance.}
OpenTelemetry defines common trace semantics and evolving GenAI agent/tool span
conventions \cite{otel-genai}.  W3C PROV supplies a general vocabulary for
entities, activities, and derivation \cite{w3c-prov-dm}.  \system is a
domain-specific evidence layer over such telemetry concepts: it defines Skill
identity, progressive-load boundaries, evidence grades, and versioned adapter
capability. It can import supported trace exports and export normalized evidence
through OTLP/HTTP without reducing a Skill run to a model or tool span.

\paragraph{Agent Skills.}
The open Agent Skills specification standardizes directory structure,
\texttt{SKILL.md} metadata, and progressive disclosure
\cite{agent-skills-spec}.  Empirical work on agentic coding manifests studies
how repositories use harness-facing instruction files \cite{coding-manifests}.
SkillsBench evaluates whether packaged Skills improve task outcomes across
diverse tasks \cite{skillsbench}, while SWE-Skills-Bench targets reusable
software-engineering Skills \cite{sweskillsbench}. Skill Coverage instead
extracts behavior constraints from
Skill instructions and grades whether trajectories cover and satisfy them
\cite{skillcoverage}. That is a trajectory-level test-adequacy question; our
complementary unit is one runtime Skill occurrence under incomplete telemetry
from installed harnesses: which lifecycle boundary is observable, what evidence
supports it, and what remains unknown. Constraint coverage could consume such
runtime evidence, but \system does not estimate instruction coverage.

\section{Threats to Validity}

\paragraph{Construct validity.}
The fault conditions are authored overlays with deterministic labels.  They
cover lifecycle boundaries but cannot represent the full distribution of agent
or Skill failures.  Exact diagnosis measures agreement with this contract, not
general semantic understanding.

\paragraph{Internal validity.}
Agent and model services may change. We freeze available versions and metadata;
comparisons remain version-specific. One run per external-validity cell
precludes variance estimates for Agent behavior. Failed
response and availability gates are retained rather than selectively retried;
prompts, schemas, and unavailable provider revisions remain part of the
diagnostic construct.

\paragraph{External validity.}
Six repositories, three agents, two model interfaces, and 19 de-identified
traces exceed a single-harness fixture but do not represent all Skills or
agents. The traces come from one local database; one Linux environment shares
a host with macOS, and only one remote PAI-DSW host is used. Authenticated
remote service paths are integration-tested, but the study does not estimate
multi-host load, availability, or multi-tenant isolation.

\paragraph{Conclusion validity.}
Claims are descriptive and mechanism-level: we neither infer causal Skill
effectiveness from one run nor pool a backend that fails its completion gate.
The 126/126 graph result is expected for preregistered relations, not
novel-fault accuracy. The 126 rows instantiate seven templates with clustered
predictions, so RQ2 reports template-stratified directions and descriptive
counts rather than case-level significance.

\section{Privacy, Ethics, and Artifact Practice}

The product defaults to local, non-intervening collection. Authenticated
self-hosting keeps viewer and Collector roles separate; observability export is
independently opt-in. Research exports omit raw prompts, code, paths, payloads,
timestamps, credentials, and identities. Opaque IDs preserve within-case
relations without source locators; raw, normalized, and inferred records remain
separate. The manifest binds fixtures, reports, and verification outputs by
digest. Release retains these safeguards and each source repository's licensing
constraints.

The system is diagnostic, not a security gate.  It does not block agent actions
or claim that missing evidence proves malicious behavior.  Model-generated
diagnoses remain \inferred and must expose unavailable or malformed states.

\paragraph{Artifact.}
Public \href{https://github.com/hellogxp/skill-runtime-intelligence}{source} and the
\href{https://github.com/hellogxp/skill-runtime-intelligence-supplementary/releases/tag/profes-2026-submission-v1}{frozen artifact}
map tables to scripts, fixtures, and JSON reports.
At its clean public source snapshot, \texttt{python -m pytest tests -q} passes
279 tests with 10 environment-dependent skips and four subtests; the selected
suite passes 13/13 gates. A versioned, de-identified,
\href{https://github.com/hellogxp/skill-runtime-intelligence/releases/tag/profes-2026-artifact-v2}{data release}
preserves failed gates, pilots, the PAI-DSW environment, digests, and
dirty-worktree provenance. Withheld source identities and unavailable provider
revisions permit protocol/output audit and deterministic verification, not
bitwise regeneration of live Agent or model calls.

\section{Conclusion}

\system makes the Agent Skill runtime visible beyond session, model-call, and
tool traces through a passive, evidence-graded lifecycle. Across 126 controlled
executions, sessions are exactly correlatable but adapter semantics and
diagnostic signatures diverge. Known-rule graph conformance, model
incompleteness, and clean-case failure-status false positives justify keeping
reproducible facts authoritative and probabilistic assistance subordinate. The
design rules are to preserve source evidence, expose unknowns, qualify adapters
as measurement instruments, separate outcomes from telemetry, and reserve
causal language for experiments that support it.

\bibliographystyle{splncs04}
\bibliography{references}

\end{document}